# IMPLEMENTING A TEST STRATEGY FOR AN ADVANCED VIDEO ACQUISITION AND PROCESSING ARCHITECTURE

Radu ARSINTE (*)

(*) *Technical University Cluj-Napoca , Tel: +40-264-595699, Str. Baritiu 26-28, Radu.Arsinte@com.utcluj.ro*

**Abstract:** This paper presents some aspects related to test process of an advanced video system used in remote IP surveillance. The system is based on a Pentium compatible architecture using the industrial standard PC104+. First the overall architecture of the system is presented, involving both hardware or software aspects. The acquisition board which is developed in a special, nonstandard architecture, is also briefly presented. The main purpose of this research was to set a coherent set of procedures in order to test all the aspects of the video acquisition board. To accomplish this, it was necessary to set-up a procedure in two steps: stand alone video board test (functional test) and an in-system test procedure verifying the compatibility with both OS: Linux and Windows. The paper presents also the results obtained using this procedure.

## I. SYSTEM ARCHITECTURE

Originally video surveillance was done based on analog technology - closed circuit television (CCTV) - and recording on video tapes. This was fine for recording what was going on, but it didn't broadcast actual live information, so it wasn't practical for monitoring stores, for instance, from a remote location. It simply provided what happened after the fact. The picture quality was low and it relied on human reliability as well – to watch and to change the tapes regularly. With the Internet revolution and the ever-increasing presence of Local Area Networks, technology has great advances in video surveillance in the 1990's. Analog camera tubes were replaced with CCD (Charged Coupled Devices) and digital cameras became affordable for most people.

With IP-based video surveillance, it is possible to connect the surveillance camera or cameras to any network or wireless adapter, and this makes extremely flexible the placement of the camera itself. A typical PC-attached video camera, while providing digital picture image quality, still has to be within approximately ten feet of the computer itself.

Using Embedded IP based video surveillance makes possible additional functions:

- IP-based recording means instant transmittal of images anywhere in the world.
- Can monitor multiple cameras from one remote location
- No decrease in recording quality over time or with repeated replays
- Digital picture quality far superior to analog
- IP-base recording is highly compressed for easier storage and can be transported over a variety of media
- Digital images can be encrypted for security purposes
- Updates and add-ons are relatively inexpensive through software packages and Internet computer networking
- Adjustable frame rates
- Remote or shared viewing may be done over the Internet or a wireless connection
- Standard IP video compression techniques could be used
- IP surveillance cameras may be added individually or in groups according to your needs

The Embedded IP surveillance system that benefits from the test procedure described in this paper has roughly the following architecture (Fig.1 [1]).

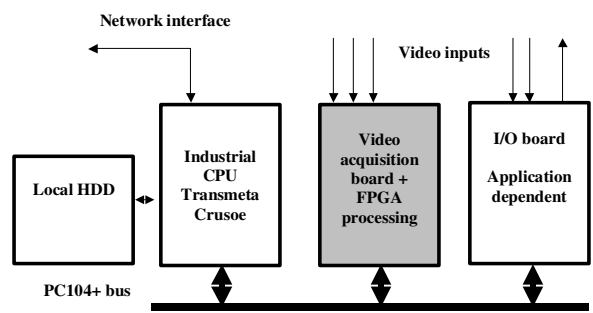

*Fig.1. IP Surveillance system architecture*

In this drawing the test targeted VideoFPGA board is dashed. The video acquisition board has a nonstandard architecture, adding along the video acquisition and MPEG encoding features, a FPGA core performing some video processing specific tasks. This makes possible to implement intensive video processing applications into FPGA and let the CPU to perform concurrently additional tasks.

The simplified architecture of the board is presented in the following image (Fig.2).

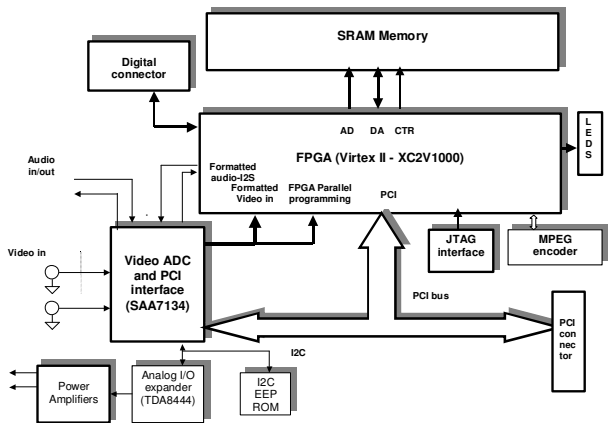

*Fig.2. VideoFPGA board architecture*

Without entering into technical details, which are not the purpose of this paper, we will present the software architecture of the application (Fig.3), also with a large degree of simplification.

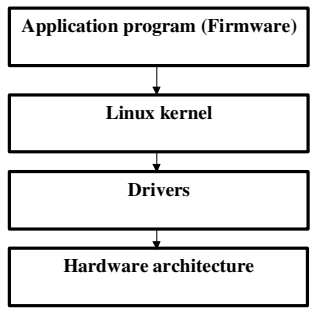

*Fig.3. Software architecture of IP surveillance system*

## II. FUNCTIONAL TEST PROCEDURE

This test is performed in a normal PC (using an PC104+ to PCI adapter) or in the embedded PC104+ system (using Windows mode).
The verification procedure of board identification has the following points:
- startup of PC in Windows mode
- observing during boot process the PCI devices listing where the correctly identified board appears ([2])
- In Device Manager (Sound, Video and Game Controllers) the board (Philips SAA7134) should appear like in the following picture (without ! mark).
- If this example (Fig.4) the (!) mark indicates an error in driver compatibility
- The driver used in this phase is a standard Hybrid driver, used practically by all the manufacturers of TV Tuner boards (Europa board driver- [3],[4]).

**Internal EEPROM initialization**
This initialization is done using a special utility developed by Philips. This utility called URD (Universal Register Debugger [5]) is based on a special script language allowing to develop complex test programs using I2C bus protocol. For this application the application file UPCB1B.urd was developed during this work and is used to test and initialize the EEPROM content.

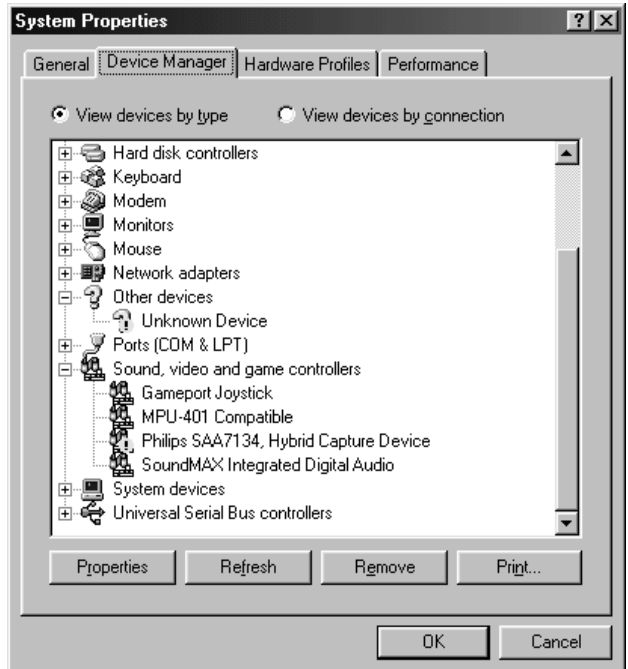

*Fig.4. Identification of the board in a Windows compatible system*

After loading the utility and application file, it is possible to display the actual content of the board's setup EEPROM. The window for an un-initialized board could look like the following image (Fig.5).

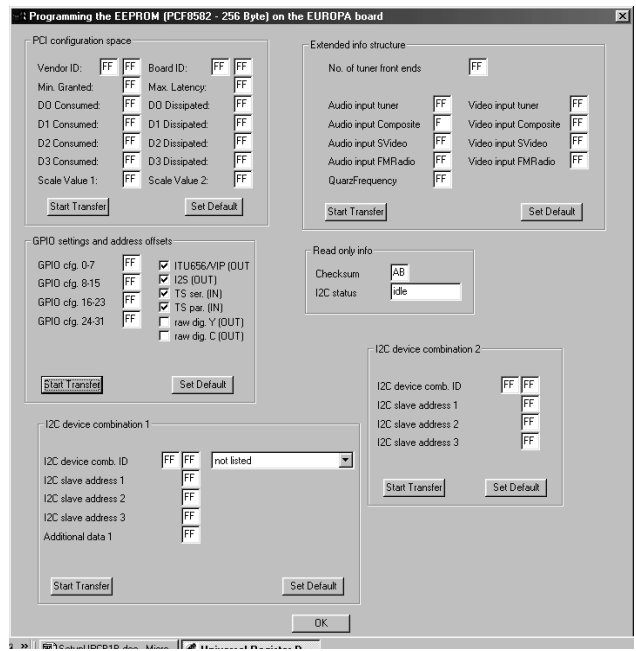

*Fig.5. Un-initialized EEPROM content of VideoFPGA board*

Filling in the content with the appropriate values for the board (equipped either with XC2V250 or XC2V1000 FPGA's) allows recognition and use of the board in system.
The following image explains the memory map for the two different configurations.
The content for XC2V250 board version is presented in fig.6.

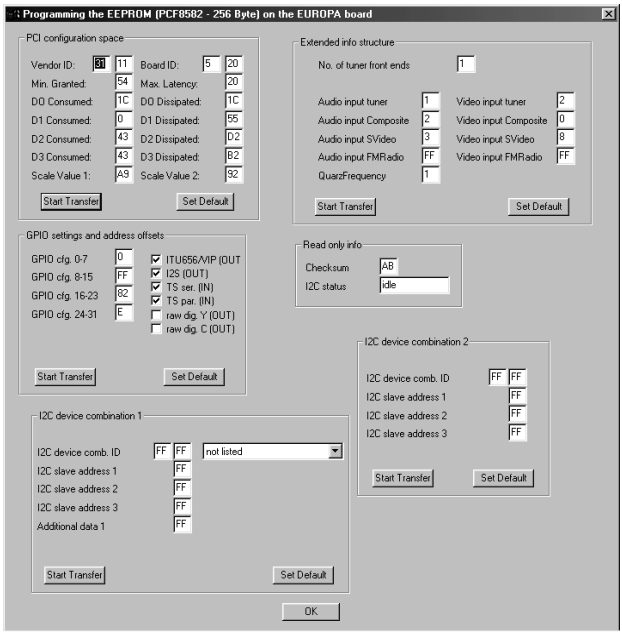

*Fig.6. EEPROM content of VideoFPGA board*

The content for XC2V1000 board version is similar excepting the third byte (04). This modification is used in the driver to identify the board type and load the correct FPGA configuration file. The rest of the content is practically imposed by the Philips SAA7134 Hybrid driver.

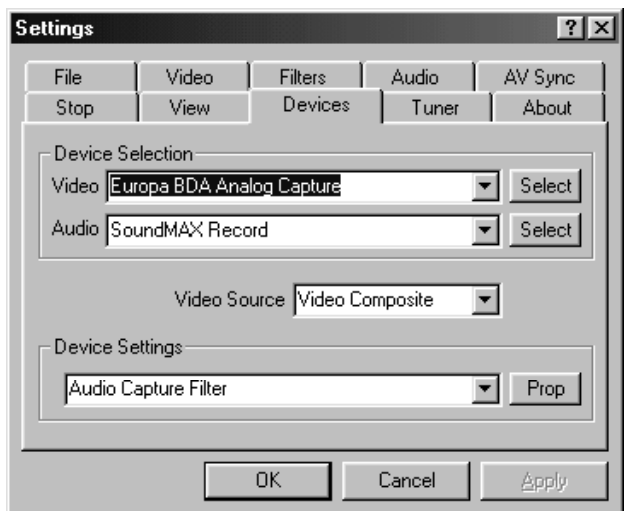

*Fig.7. Selecting the BDA driver and video input*

**Test of video streaming with live video viewing program**
To perform this test the video source (pattern generator or other video source, video tuner for example) was applied to one of the video inputs (Vid0 or Vid1).
This test can be done with different programs. For example Dscaler, VirtualVCR, or Amcap (delivered by Microsoft in DirectX package). Depending on the configuration (video board, motherboard) one or all the programs are functional.
The next description applies to VirtualVCR application.
Selecting "Europa BDA analog capture" in Devices menu with video source "Video Composite" for input "Vid1" or "Video Tuner" for "Vid0" (Fig.7) makes possible to visualize the video stream generated by SAA7134. In few moments the live video appears on the window, like in Fig.8. This marks the fact that the board is correctly installed, the SAA7134 IC is on place and is working correctly. This makes possible to start the final "In system" test procedure (see chapter III).
The same result could be obtained using DirectX prototyping technology. In this approach the components are assembled using GraphEdit utility from DirectX SDK ([6]) . An example of test graph for this board is presented in Fig.9.

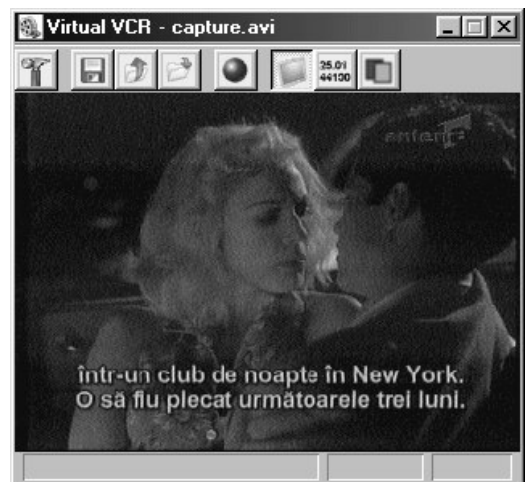

*Fig.8. Live video activation*

### III. IN SYSTEM TEST PROCEDURE
The minimal test system is composed from a server (PC104

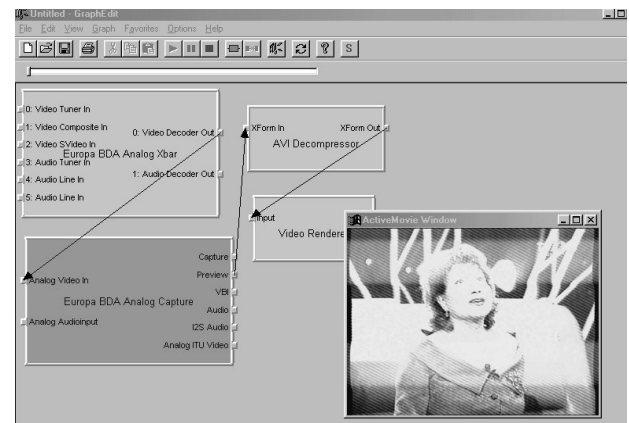

*Fig. 9 Video streaming debugging using GraphEdit*

system with the acquisition board) and a client (normal PC). The proposed configurations are presented in Fig.10 (a and b). If this test system (involving a local network and a DHCP server – Fig.10.a) is not possible to be implemented, it is possible to use a simplified version based on a direct connection using a UTP crossover cable (Fig.10.b).

In this case the IP address assumed in the following procedure

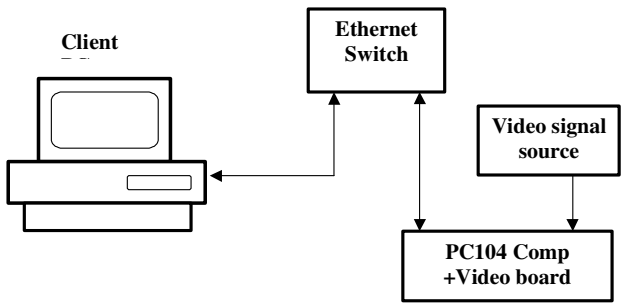

*Fig.10.a Test connection using a switch*

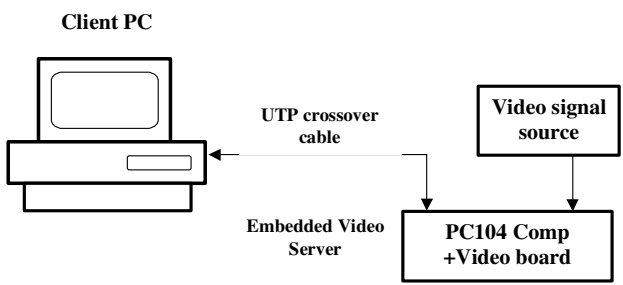

*Fig.10.b Test connection using a direct cable*

(192.168.0.200) should be replaced with the default address 10.1.1.1 allocated at startup by Linux init procedure.
Preliminary operations necessary to apply this procedure:
- Installation of Mozilla Firefox browser in Client PC
- Connection of the client and the server directly or via Ethernet switch

Main Steps of testing
- Startup of PC104 system in Linux mode (this is the default option
- Start of browser and accessing the http://192.168.0.200/videofpga.html

This should open the main test server page as in Fig.11. From this window it is possible to launch individual tests, for different functional blocks.

**Image grabbing test**
"Grab image" will create in the left window after few seconds an image with the captured frame (Fig.12).

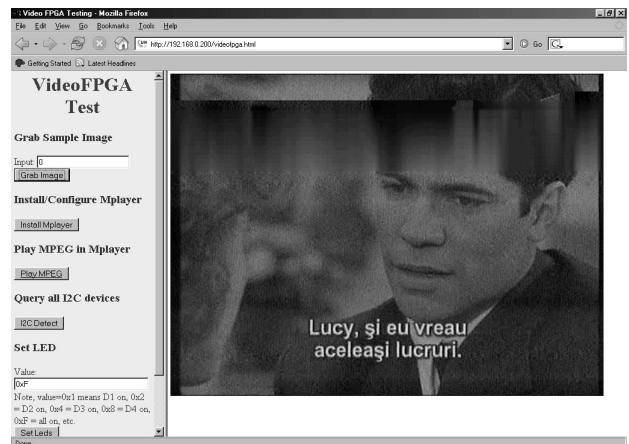

*Fig.12. Display of image grabbing test*

**I2C scan and detection**
This function verifies the on-board I2C addresses, identifying the I2C devices present in the system.
The result is presented in Fig.13.

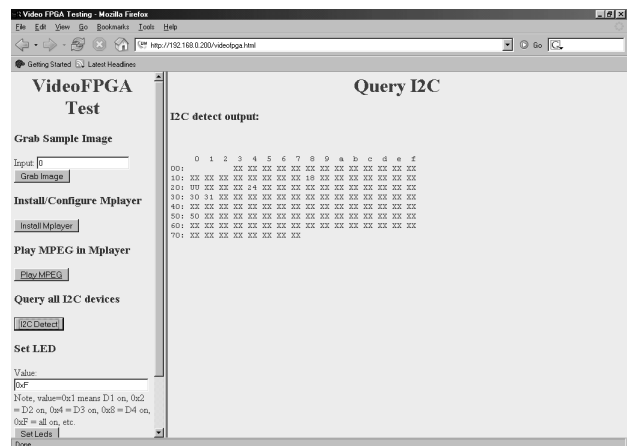

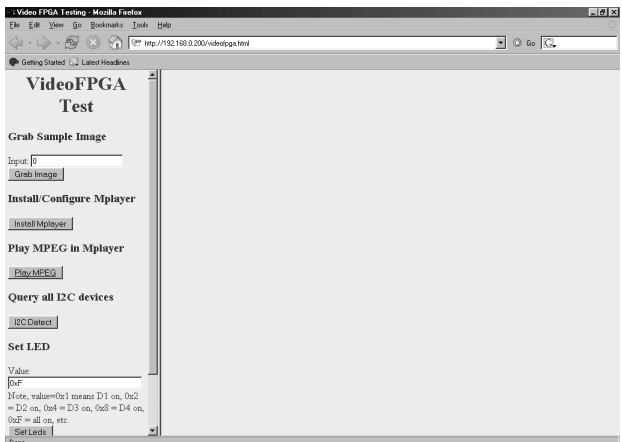

*Fig.11. Main page of test server*

*Fig.13. I2C devices identification*

The I2C device list should contain all the addresses for the I2C compatible devices on board. In this example the identified addresses are:
18 – LM83, 20- MPEG encoder, 24 – TDA8444, 30/31 FPGA I2C core, 50 – EEPROM

**Control LED test**
This function should allow global or individual LED lighting (Fig.14).

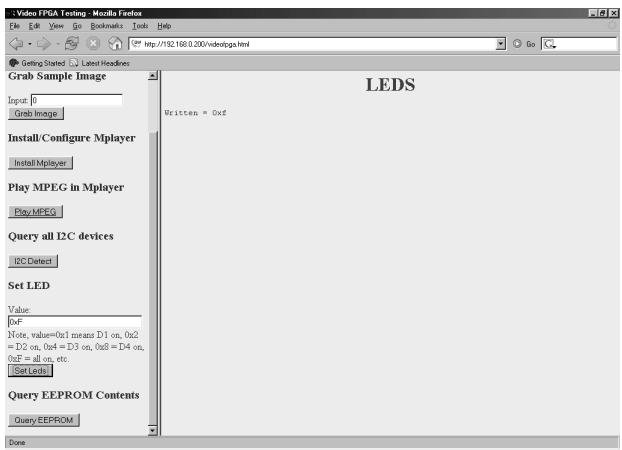

*Fig.14. LED test screen*

**EEPROM content verification**
"Query EEPROM" button will dump in the right window a EEPROM content dump (Fig.15).

**Streaming video test**

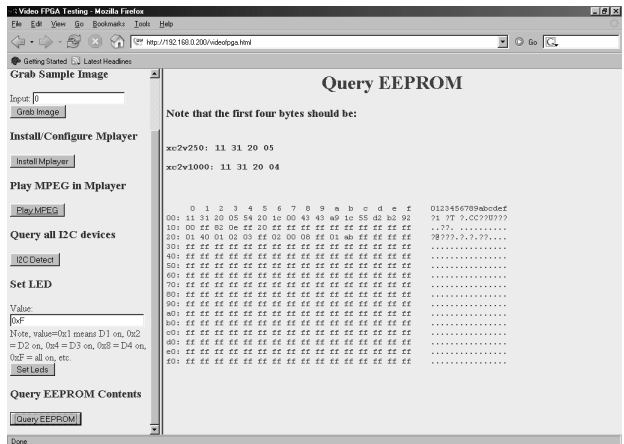

*Fig.15. EEPROM query test*

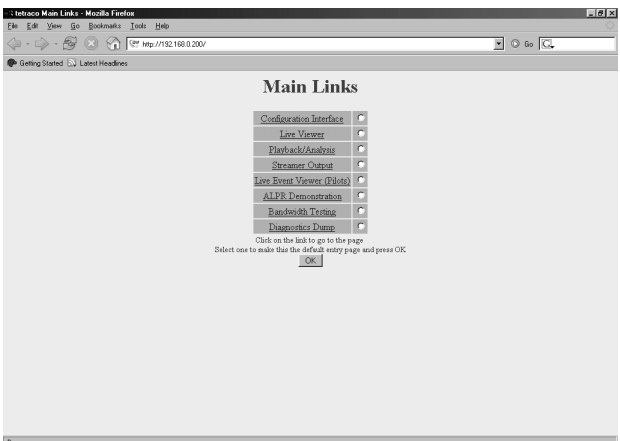

*Fig.16. Main page of video server*

Opening http://192.168.0.200/, main page of video server will create the following menu (Fig.16).
Streamer Output link will create a screen where All live cams link creates "near" live video (moving images) on your screen.

## IV. CONCLUSIONS

This "simple" and affordable procedure allows the full functional test of video streaming and board integration in the overall architecture. It is based entirely on tested and reliable technologies (DirectX, Windows, Linux drivers) proofing that is not necessary to write standalone dedicated programs, which could take many months of work. All the programs (excepting the final procedure, which is extracted from the end application) are free of charge and easy to implement and understand.

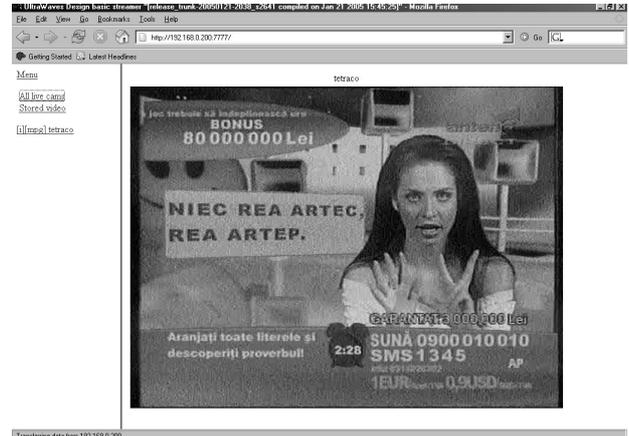

*Fig.17. Live video streaming from video server*

**Future work**
- Extending the test pattern for other resources of the board (not included in this test procedure). This applies for example for FPGA memory.
- Building specific drivers for this board eliminating the features of SAA7134 unused in this application, to accelerate the video streaming and minimizing memory allocation.
- Extending the results for other video acquisition architectures based, for example, on Conexant video chips

**REFERENCES**
[1] R. Arsinte - Video FPGA board specification – Technical report, Telar Tech, 2004
[2] * * *, Philips Tuner Driver Development Kit -User Manual, 2003
[3] * * *, SAA7134HL -PCI audio and video broadcast decoder, Product specification, 2002
[4] * * *, SAA7130; SAA7134 PCI audio and video broadcast decoder, User manual, 2002
[5] * * *, User Manual for the Universal Register Debugger (URD), Philips Semiconductor, 2001
[6] * * *, Philips SAA713x SDK User Manual, 2003*Originally published in Acta Technica Napocensis Journal ISSN 1221-6542*
*Technical University of Cluj-Napoca & Mediamira Publishing 2006*

5